\pdfoutput=1
\documentclass[a4paper,twoside]{article}
\usepackage{rfia2000}
\usepackage{times}
\usepackage{framed,multirow}

\usepackage{amssymb}
\usepackage{amsmath}
\usepackage{latexsym}
\usepackage{makecell}
\usepackage{tikz}
\usetikzlibrary{arrows}
\usepackage{subfigure}

\usepackage[linesnumbered,ruled]{algorithm2e}
\usepackage{booktabs}

\usepackage{url}
\usepackage{xcolor}
\definecolor{newcolor}{rgb}{.8,.349,.1}

\newtheorem{definition}{Definition}

\tikzstyle{block} = [draw, rectangle, 
minimum height=0.7em, minimum width=6em, align=center,inner sep=0,outer sep=0,node distance=0.6em]
\tikzstyle{pooling} = [align=center, outer sep=0cm,inner sep=0cm,node distance=0.4cm]
\tikzstyle{afterpool} = [outer sep=0cm, node distance=0.3cm]
\begin{document}

\date{}
\title{A Convolutional Neural Network into graph space}
\author{\begin{tabular}[t]{c@{\extracolsep{4em}}c@{\extracolsep{4em}}c@{\extracolsep{4em}}c}
Chloé Martineau${}^1$ & Romain Raveaux${}^1$ & Donatello Conte${}^1$ & Gilles Venturini${}^1$ \\
\end{tabular}
{} \\
 \\
${}^1$        Laboratoire d'Informatique Fondamentale et Appliquée de Tours (LIFAT - EA 6300)
\\
Université François Rabelais
{} \\
 \\
64, avenue Jean Portalis, 37~200~Tours, France \\
\{firstname.lastname\}@univ-tours.fr
}
\maketitle

Under consideration at Pattern Recognition Letters

\subsection*{Abstract}
{\em
Convolutional neural networks (CNNs), in a few decades, have outperformed the existing state of the art methods in classification context. However, in the way they were formalised, CNNs are bound to operate on euclidean spaces. Indeed, convolution is a signal operation that are defined on euclidean spaces. This has restricted deep learning main use to euclidean-defined data such as sound or image.

And yet, numerous computer application fields (among which network analysis, computational social science, chemo-informatics or computer graphics) induce non-euclideanly defined data such as graphs, networks or manifolds.

In this paper we propose a new convolution neural network architecture, defined directly into graph space. Convolution and pooling operators are defined in graph domain. We show its usability in a back-propagation context.

Experimental results show that our model performance is at state of the art level on simple tasks. It shows robustness with respect to graph domain changes and improvement with respect to other euclidean and non-euclidean convolutional architectures.
}

\graphicspath{{./images/}}

\section{Introduction}
\label{sec:introduction}

Graphs are frequently used in various fields of computer science, since they constitute a universal modeling tool which allows the description of structured data. The handled objects and their relations are described in a single and human-readable formalism. Hence, tools for graphs supervised classification and graph mining are required in many applications such as pattern recognition \cite{livreRiesen15}, chemical components analysis \cite{gauzere}, structured data retrieval \cite{DBLP:journals/jvcir/RaveauxBO13}. 
\subsection{Graph Classification}
Graph classifiers can be categorized into two categories whether the classifier operates in a graph space or in a vector space.

\subsubsection{Graph space}
Graph space classification consists of finding a metric $d: \mathbb{G} \times \mathbb{G} \rightarrow \mathbb{R})$ (with $\mathbb{G}$ the graph space) to evaluate the dissimilarity between two graphs. This metric can be later used in a K-Nearest Neighbor context, where the distances between the object to be classified and the elements in the learning database are used as a base for classification.  The similarity or dissimilarity between two graphs requires the computation and the evaluation of the "best" matching between them. Since exact isomorphism rarely occurs in pattern analysis applications, the matching process must be error-tolerant, i.e., it must tolerate differences on the topology and/or its labeling. For instance, in the Graph Edit Distance (GED) problem \cite{livreRiesen15}, the graph matching process and the dissimilarity computation are linked through the introduction of a set of graph edit operations. Each edit operation is characterized by a cost, and the dissimilarity measure is the total cost of the least expensive set of operations that transform one graph into another one.
In \cite{livreRiesen15,DBLP:journals/prl/BougleuxBCFGV17}, the GED is shown to be equivalent to a Quadratic Assignment Problem (QAP). Since error-tolerant graph matching problems are NP-hard most research has long focused on developing accurate and efficient approximate algorithms.
In \cite{DBLP:journals/prl/BougleuxBCFGV17}, with this quadratic formulation, two well known graph matching methods called Integer Projected Fixed Point method \cite{Leordeanu2009} and Graduated Non Convexity and Concavity Procedure \cite{LiuQ14} are applied to GED. In \cite{Leordeanu2009}, this heuristic improves an initial solution by solving a linear assignment problem (LSAP) and a relaxed QAP where binary constraints are relaxed to the continuous domain. The algorithm iterates through gradient descent using the Hungarian algorithm to solve the LSAP  and a line search. In \cite{LiuQ14}, a path following algorithm aims at approximating the solution of a QAP by considering a convex-concave relaxation through a modified quadratic function.

\subsubsection{Vector space}
Vector space graph classification is about representing graphs as vectors to classify them.

A first one consists in transforming the initial structural problem in a common statistical pattern recognition one by describing the graphs with vectors in an Euclidean space \cite{DBLP:journals/pr/LuqmanRLB13}.
In such a context, some features (vertex degree, labels occurrence histograms,etc.) are extracted from the graph. Hence, the graph is projected in a Euclidean space and classical machine learning algorithms can be applied. Such approaches suffer from a main drawback: to have a satisfactory description of topological structure and graph content, the number of such features has to be very large and dimensionality issues occur.

Another possible approach also consists in projecting the graphs in a Euclidean space of a given dimension but using a distance matrix between each pairs of graphs. In such cases, a dissimilarity measure between graphs has to be designed \cite{DBLP:conf/sspr/BunkeR08}. Kernels can be derived from the distance matrix. It is the case for multidimensional scaling methods proposed in \cite{DBLP:journals/pami/RothLKB03}.

Alternatively, graph embedding can be implemented implicitly through kernel-based machine learning algorithms. In the kernel approaches, an explicit data representation is of secondary interest. That is, rather than defining individual representations for each pattern or object, the data at hand is represented by pairwise comparisons only. The graphs are not explicitly but implicitly projected in a Euclidean space without defining the function $\phi$. More formally, under given conditions, a similarity function can be replaced by a graph kernel function  $k : <\mathbb{G}, \mathbb{G}> \rightarrow \mathbb{R} $. Most kernel methods can only process kernel values which are established by symmetric and positive definite kernel functions.
Many kernels have been proposed in the literature \cite{DBLP:series/smpai/NeuhausB07,gauzere}. In most cases, the graph is embedded in a feature space composed of label sequences through a graph traversal. According to this traversal, the kernel value is then computed by measuring similarity between label sequences. Even if such approaches have proven to achieve high performance, they suffer from their lack of interpretability. In fact, it is very difficult to come back to graph space from the kernel space. This problem is also known as "pre-image". 
\subsection{Euclidean and geometric deep learning}
Deep learning has achieved a remarkable performance breakthrough in several fields, most notably in speech recognition, natural language processing, and computer vision.
In particular, convolutional neural network (CNN) architectures currently produce state-of-the-art performance on a variety of image analysis tasks such as object detection and recognition. Most of deep learning research has so far focused on dealing with 1D, 2D, or 3D Euclidean structured data such as acoustic signals, images, or videos.

Recently, there has been an increasing interest in geometric deep learning, attempting to generalize deep learning methods to non-Euclidean structured data such as graphs and manifolds, with a variety of applications from the domains of network analysis, computational social science, or computer graphics. 
Graph neural networks are one of possible ways to implement explicit graph embedding: the neural network takes a graph as input and outputs a vector. This one can be used for classification. Moreover, graph neural networks perform learning the explicit embedding according to a given learning criterion.

\subsection{Graph Neural Networks}
These neural networks often try to apply convolution to graphs so that it mimics classical convolutional neural networks. Convolution definition on graph space is a tedious theoretical task. There is indeed no straightforward definition. However, one can identify two families of definitions in the existing literature. The first family (spectral approaches) relies on the convolution theorem. This theorem states that the convolution operator on the spatial domain is equivalent to the product operator on the frequency domain. Although this theorem was only proven on euclidean spaces, a group of approaches in the litterature postulates its validity on the graph space. A graph frequency domain is accessed through diagonalization of its Laplacian $L = D - A$ ($D$ and $A$ respectively being the degree and adjacency matrices of the graph).
Such approaches have two main limitations. The first one is their sensitivity to topological variations: a slight deformation of the graph structure changes the resulting convolution signal drastically. The latter is that there is no Fast Fourier Transform on the graph space: as previously stated, accessing the graph frequency domain relies on matrix diagonalization and therefore inversion. Inverting a matrix is a costly operation.

These drawbacks exist because convolution is applied implicitly to the graph through its frequency domain. A simple way to avoid them is to apply convolution directly on the spatial domain. The second family of approaches (the spatial ones) try to come up with analogies of the original convolution definition. However, existing approaches often degrade graphs and therefore do not fully exploit their structural information.

In this paper, we propose a graph convolution operator which operates solely on graph space. This is made possible through usage of graph matching to define local convolutional operation.
By doing so, we try to establish a link between two scientific communities who respectively work on graphs and deep learning. More specifically, we define graph-based computations using operators from the graph matching litterature in a deep learning (neural network) framework.

\section{State of the Art}
\label{sec:stateoftheart}
This section offers a review of existing graph neural network definitions.
Every graph neural network layer can then be written as a non-linear function:
$$H^{(l+1)} = f(H^{(l)}, A) $$

As an example, let's consider the following very simple form of a layer-wise propagation rule:
$$f(H^{(l)}, A) = \sigma\left(D^{-1}AH^{(l)}W^{(l)}\right)$$
$\sigma(.)$ is a non-linear activation function like the ReLU. 

Multiplying the input with $D^{-1}A$ now corresponds to taking the average of neighboring node features from the layer $l$. 
It is also called in the literature "average neighbor messages" and it acts like passing average node features from one layer to another. 
In \cite{DBLP:journals/corr/KipfW16}, a better (symetric) normalization of the adjacency matrix is proposed i.e. $D^{-\frac{1}{2}}AD^{-\frac{1}{2}}$. A per-neighbor normalization is performed instead of simple average, normalization varies across neighbors.
$$f(H^{(l)}, A) = \sigma\left( \hat{D}^{-\frac{1}{2}}\hat{A}\hat{D}^{-\frac{1}{2}}H^{(l)}W^{(l)}\right) $$
with $\hat{A}=A+I$, where $I$ is the identity matrix and $\hat{D}$ is the diagonal node degree matrix of $\hat{A}$. 
The complexity of this model is $O(|E|)$ time complexity overall (E being the set of edges). 

More operations have been investigated in the literature \cite{DBLP:journals/corr/NowakVBB17}. A complete family of operations can be used :
\begin{itemize}
    \item \textbf{I} : this identity operator does not consider the structure of the graph and neither provide any aggregation. Used alone this operator makes the GNN a composition of  $|V|$ MLP completly independent. One MLP for each node feature vector.
    \item $A$ : the adjacency operator gather information on the node neighborhood (1 hop).
    \item $D$ : $D=diag(A\mathbf{1})$. This degree operator gather information on the node degree. $D$ is node degree matrix (a diagonal matrix).
    \item $A_j$ : $A_j = min(1, A^{2^{j}})$. It encodes $2^j$-hop neighborhoods of each node, and allow us to aggregate local information at different scales, which is useful in regular graphs. 
    \item $U$ : $U$ is matrix filled with ones. This average operator, which allows to broadcast information globally at each layer, thus giving the GNN the ability to recover average degrees, or more generally moments of local graph properties.
\end{itemize}

Let us denote $\mathcal{A} = \{\mathbf{1}, D, A, A_1, \cdots, A_J, U\}$. A GNN layer is defined as : 
$$f(H^{(l)}, \mathcal{A}) = \sigma\left( \sum_{B \in \mathcal{A}}  BH^{(l)}W_B^{(l)}\right) $$
$\Omega=\{W^{(l)}_1, \cdots, W^{(l)}_{|\mathcal{A}|} \}$, $W^{(l)}_B \in \mathbb{R}^{m_{(l)} \times m_{(l+1)}}$ are trainable parameters.

Key distinctions are in how different approaches aggregate messages. So far, proposals have aggregated the neighbor messages by taking their (weighted) average, but is it possible to do better?
In \cite{DBLP:journals/corr/HamiltonYL17}, a GNN called GraphSAGE is proposed. The aggregation of neighbors information is more complex. The very general scheme of aggregation can written thanks to the function $AGG$:
$$H^{(l+1)}= \sigma\left(AGG(H^{(l)})W^{(l)}\right)$$
Let us define $\mathcal{N}(u)$ is the set of nodes in the 1-hop neighborhood of node $u$.
\begin{itemize}
    \item mean : $AGG_u= \frac{1}{|\mathcal{N}(u)|}\sum_{v \in \mathcal{N}(u)} H^{(l)}_v \quad \forall u \in V  \implies AGG=D^{-1}AH^{(l)} $. 
    \item max : $AGG_u=max(\{ H^{(l)}_{v},  \quad \forall v \in \mathcal{N}(u) \} )  \quad \forall u \in V$ . Transform neighbor vectors into a matrix and apply a max pooling element-wise.
    \item LSTM : $AGG_u=LSTM([H^{(l)}_{v}, \quad \forall v \in \pi(\mathcal{N}(u))])  \quad \forall u \in V$. Where $\pi$ is a random permutation. The idea is to provide to the LSTM a sequence composed of neighbor embeddings. So the input sequence is composed of vectors. The sequence is randomly permuted by the function $\pi$.
\end{itemize}

In \cite{DBLP:journals/corr/MontiBMRSB16}, the graph structure is locally embedded into a vector space. The distribution of local structures in the local space is estimated by a Gaussian Mixture Model. The $AGG_u$ function is then expressed by a mixture of Gaussians. The Gaussian parameters are covariance matrix and mean vector and they are learnt during the training of the neural network. 


A notable variant of GNN is graph attention networks (GAT),
which was first proposed in \cite{2017arXiv171010903V}. This model includes the self attention mechanism to evaluate the individual importance of the adjacent nodes and therefore it can be applied to graph nodes having different degrees by specifying arbitrary weights to the neighbors \cite{2017arXiv171010903V}.

For further reading, good surveys about graph neural networks have been published \cite{DBLP:journals/corr/abs-1812-08434,DBLP:journals/corr/abs-1812-04202,DBLP:journals/corr/abs-1901-00596}.




\paragraph*{Deadlocks, contributions and motivations}
From the literature, two main deadlocks can be drawn. First, in many of the related works \cite{DBLP:journals/corr/KipfW16,DBLP:journals/corr/NowakVBB17,2017arXiv171010903V}, edge features are not well considered. However, the edge information is of first interest to boosts the structural knowledge in the computation of the node embedding. Second, most of the aforementioned approaches do not take full advantage of the graph topology \cite{DBLP:journals/corr/MontiBMRSB16,DBLP:journals/corr/KipfW16}. The graph structure is locally embedded into a vector space (i.e. the tangent space at a given point of a riemannian manifold).
In this paper, we propose CNN architectures that remain in the graph domain. Especially, we design a convolution operator onto graph space through the solution of a graph matching problem. The problem of graph matching under node and pair-wise constraints is fundamental to capture topological information. It takes into account the nodes and edge features along with their neighborhood structure. Consequently, graph matching-based convolution can release deadlocks related to edge information integration, domain changes sensitivity and Euclidean space projection. Graph matching can be seen as added local constraints in the machine learning problem. We promote a truly novel class of neural network architecture where layers contain a combinatorial optimization scheme that plays a fundamental role in the construction of the entire neural network architecture. Consequently, we highlight the interplay between machine learning and combinatorial optimization.

\section{Graph Convolutional Neural Network}
\label{sec:definitions}

\subsection{Notation}

Frequently used notations are summarized
in Table~\ref{tab:notation}.

\begin{table}
	\centering
	\caption{Frequently used notations}
	\begin{tabular}{l|l}
		Notation & Description   \\ \hline
		$G_I$ & An input graph  \\
		$G_F$ & A filter graph  \\
		$g_I^i$ & Neighbourhood subgraph rooted at vertex $i$ in $I$\\
		$i$, $j$ & Vertices in graph $G_I$\\
		$ij$ & An edge in graph $G_I$ between $i$ and $j$  \\
		$a$ & A vertex in $G_F$ \\
		$ab$ & An edge in $G_F$ between $a$ and $b$ \\
		$\mu$ & Labelling function for vertices \\
		$\zeta$ & Labelling function for edges \\
		$G_F^W$ & A filter graph and its associated weights  \\
		$\mu(a)$ & Vertex label of $a$ \\
		$\mu^W(a)$ & Vertex label of $a$ parametrized by $W$\\
		$\vert \Omega_{ij} \vert$ & Cardinality of $ \Omega_{ij} $ \\
		$\delta_x^y$ & Kronecker delta of $x$ and $y$
	\end{tabular}
	\label{tab:notation}
\end{table}

\subsection{Graph matching}
\label{subsec:graphmatching}
To define our convolution operator, we must define the graph matching function that will be pointwisely used.
\subsubsection{Graph matching problem}
\label{def:gmformulation}
\begin{subequations}
	
	Let $G_1$ and $G_2$ be attributed graphs: $G_1 = (V_1, E_1, \mu_1, \zeta_1)$ and $G_2= (V_2, E_2, \mu_2, \zeta_2)$
	\begin{equation}
	\label{eq:gm}
	\begin{aligned}
	\text{GMS}(G_1, G_2) = \underset{y} {\mathrm{max}} \quad s(G_1, G_2, y),
	\end{aligned}
	\end{equation}
	\begin{align}
	\text{subject to}\quad
	& y \in \{0,1 \}^ {n_1 n_2 }\\
	& \sum_{i=1}^{n_1} y_{i,a} = 1  \quad \forall  a \in [1,\cdots,n_2] \\
	& \sum_{a=1}^{n_2} y_{i,a} \leq 1 \quad \forall i \in [1,\cdots, n_1 ]\\
	& \vert V_1 \vert \geq \vert V_2 \vert
	\end{align}
\end{subequations}

The similarity function $s$ is defined as follows:

\begin{subequations}
	\begin{equation}
	s(G_1,G_2,y) =	\sum_{y_{ia}=1} s_V(i, a)  + \sum_{y_{ia}=1} \sum_{y_{jb}=1} s_E(ij, ab)
	\end{equation}
	\begin{equation}
	s_V(i, a) = \mu_1(i) . \mu_2(a)
	\end{equation}
	\begin{equation}
	s_V(i, \epsilon) = s_V(\epsilon, a) = 0
	\end{equation}
	\begin{equation}
	s_E(ij, ab) = \zeta_1(ij) . \zeta_2(ab)
	\end{equation}
	\begin{equation}
	s_E(ij, \epsilon\epsilon) = s_E(\epsilon\epsilon, ab) = 0
	\end{equation}
\end{subequations}

\begin{subequations}
	Let $\pi(G_1, G_2, e)$ denote an assignment of element (edge or vertex) $e$ $\in$ $V_1 \cup E_1$ to some element in $V_2 \cup E_2 \cup \{\epsilon, \epsilon\epsilon\}$:
	
	\begin{equation}
	\pi(G_1, G_2, i)=a \iff \exists a \in V_2 : y_{ia} = 1
	\end{equation}
	\begin{equation}
	\pi(G_1, G_2, i) = \epsilon \iff \forall a \in V_2 : y_{ia} = 0
	\end{equation}
	\begin{equation}
	\pi(G_1, G_2,ij)=ab \iff \exists ab \in E_2 : y_{ia} = 1 \land y_{jb} = 1
	\end{equation}
	\begin{equation}
	\pi(G_1, G_2, ij) = \epsilon \epsilon \iff \forall a,b \in V_2 : y_{ia} = 0 \lor y_{jb} = 0
	\end{equation}
\end{subequations}

The similarity function can be rewritten as follows:

\begin{subequations}
	\begin{align}
	s(G_1,G_2,y) =&	\sum_{i \in V_1} s_V(i, \pi(G_1, G_2, i))\\
	 &+ \sum_{ij \in E_1} s_E(ij, \pi(G_1, G_2, ij))
	\end{align}
\end{subequations}

\subsection{Graph convolution based on graph matching}
Now that our matching operator is formulated, we can apply it over an input graph to compute the result of a convolution.

Let $G_I$ and $G_F$ be attributed graphs: $G_I = (V_I, E_I, \mu_I, \zeta_I)$ and $G_F= (V_F, E_F, \mu_F, \zeta_F)$. $G_I$ and $G_F$ are respectively referred to as the input graph and the filter graph.

\subsubsection{Graph convolution operator $\odot$}
\label{def:graphconv}
The graph convolution operator is a function $\mathbb{G} \times \mathbb{G} \to \mathbb{G}$ and is defined as follows:
\begin{subequations}
	\begin{equation}
	\label{eq:graphconv}
	\begin{aligned}
	G_I \odot G_F = (V_I, E_I, \mu, \zeta)
	\end{aligned}
	\end{equation}
	\begin{align}
	\text{with}\quad
	& \mu: V_I \to \mathbb{R}\quad\text{such that}\quad\mu(i) = \text{GMS}(g_I^i, G_F)\\
	& \zeta: E_I \to \mathbb{R}\quad\text{such that}\quad\zeta(ij) = \text{score}(ij, G_I, G_F)
	\end{align}
\end{subequations}
where $g_I^i$ and $score$ are defined as follows.

\subsubsection{Vertex neighbourhood graph ($l$-hops)}
$g_I^i$ is defining the neighbourhood (which is a subgraph) for vertex $i$ in $G_I$:

\begin{subequations}
	\begin{equation}
	g_I^i = (N_I^l[i], E_I^i, \mu_I, \zeta_I)
	\end{equation}
	\begin{align}
	\text{with}\quad
	& N_I^l[i] \text{ the } l \text{-hops closed neighbourhood of }i\text{ in } G_I\\
	\text{and}\quad
	& E_I^i = \{kl \in E_I \quad\text{s.t.}\quad k, l \in N_I^l[i]\}
	\end{align}
\end{subequations}

\subsubsection{Edge attribute in convolved graph}

$\text{score}$ is a function mapping an edge to its matching score in the found GMS. The problem is that it might be assigned multiple times:

\begin{equation}
\text{let}\quad\Omega_{ij} = \{g_I^k\quad \forall k \in V_I : ij \in g_I^k\} \quad\forall ij \in E_I
\end{equation} 
$\Omega_{ij}$ potentially contains more than one element. Therefore, score can be defined as follows:

\begin{subequations}
	\begin{equation}
	\text{score}(ij, G_I, G_F) = \theta \left( \{s_E(ij, \pi(g_I, G_F, ij)) \quad \forall g_I \in \Omega_{ij}\} \right)
	\end{equation}
	\begin{align}
	\text{with}\quad
	& \theta : \text{some statistical estimator (max or avg)} 
	\end{align}
\end{subequations}

\subsection{Convolution layer}
Now that the convolution operator is defined, it is possible to use it as a base to build a convolution layer. This layer can be included in a graph neural network.

\subsubsection{Graph convolution filter: the filter graph}

A graph convolution filter is an attributed graph $G_F^W$. Its role is analogous to that of a vanilla CNN kernel: it modifies the output and gets modified through backpropagation.
Every attribute function is parametrized with respect to a weight vector $W \in \mathbb{R}^{|V|+|E|}$.

\begin{subequations}
	\begin{equation}
	G_F^W = (V_F, E_F, \mu_F^W, \zeta_F^W)
	\end{equation}	
	\begin{align}
	\text{with}\quad
	& \mu_F^W(a) = W_a\\
	& \zeta_F^W(ab) = W_{ab}
	\end{align}
\end{subequations}


\subsubsection{Graph convolution layer}
A convolution layer is a set of convolution filters $\{G_F^p\}_{1 \le p \le n}$ 
applied on a same input graph $G_I$. The output of the layer consists of all filters results (analogous to euclidean convolution feature map) stacked up.

Let $u$ be the output function of the layer s.t.:
\begin{subequations}
	\begin{equation}
	u : \mathbb{G} \to \mathbb{G}\quad u(G_I) = \psi(\{u_p(G_I)\}_{1 \le p \le n})
	\end{equation}
	\begin{align}
	\text{with}\quad
	& \psi : \mathbb{G}^n \to \mathbb{G}\quad \psi(\{u_p(G_I)\}_{1 \le p \le n}) = (V_I, E_I, M, Z)\\
	& M : V_I \to \mathbb{R}^n \quad (M(i))^p = \mu_{F}^p(i)\\
	& Z : E_I \to \mathbb{R}^n \quad (Z(ij))^p = \zeta_{F}^p(ij)\\
	& n\text{ the number of filters}
	\end{align}
	
	$\psi$ function keeps only a single graph structure and concatenates each vertex/edge attribute. The output function of the layer is a graph with same topology as $G_I$ but with attributes as vectors composed by attributes of every filters outputs. \end{subequations}

\begin{figure}
	\centering
	\includegraphics[width=0.9\linewidth]{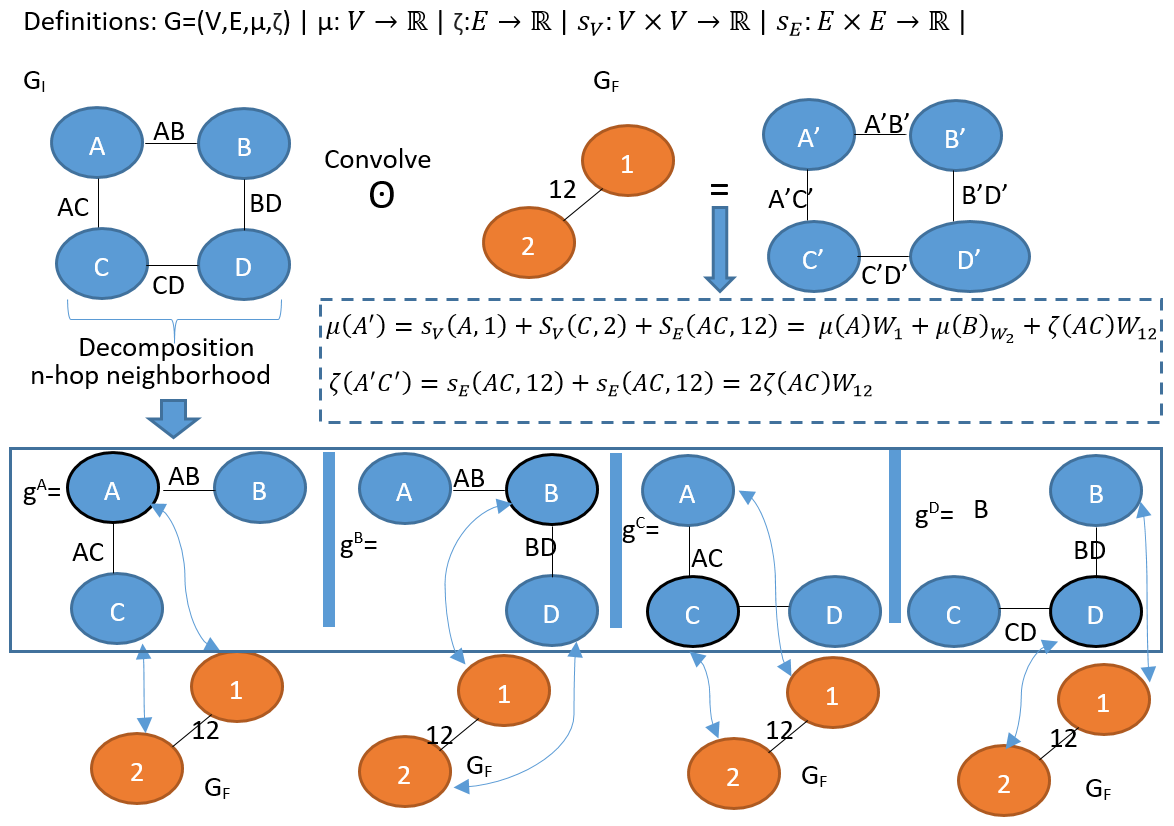}
	\caption{Computing graph convolution}
	\label{gcnn:fig:gconv}
\end{figure}

Graph convolution computation can be seen as a step-by-step process (shown in Figure~\ref{gcnn:fig:gconv}). The first step is neighbourhood extraction: for each vertice $i$ in $G_I$ (the input graph), the neighbourhood graph $g^i$ is extracted. It is composed of every neighbour of $i$ in a given range (it can be 1-hop away but also n-hops away). $g_i$ and $G_F$ (the filter graph) are matched. The matching score $\text{GMS}(g^i, G_F)$ becomes the output of the convolution at $i$.

\begin{definition}{Graph convolution differentiation}
	\label{def:gcnn:differentiation}
	Let $G_I = (V_I, E_I, \mu_I, \zeta_I)$ the input graph and $u(G_I)$ be the output of a convolution layer (called Conv) s.t. $u(G_I) = G_I \odot G_F^W$.
	
	\begin{subequations}
		\begin{align}
		u(G_I) &= G_I \odot G_F^W\\
		&= (V_I, E_I, \mu, \zeta)
		\end{align}
	\end{subequations}
	
	To simplify notations, let's consider the output of the Conv layer to be the vertex and edge labelling functions $\mu$ and $\zeta$, as neither the vertices or edges sets change during convolution. The output will be noted as in Equation~\ref{gcnn:eq:outp}
	
	Let $\mathcal{J}$ be a loss function (for example mean-squared error or categorical cross-entropy). Let's suppose Conv is involved in the calculation of $\mathcal{J}$ such that:
	
	\begin{equation*}
	\mathcal{J} = A \circ \text{Conv} \circ B
	\end{equation*}
	
	$B$ and $A$ respectively being the processing before and after Conv. In order to minimize $\mathcal{J}$, its gradient must be calculated with respect to $W$. This gradient will then be used to modify $W$ itself. For calculus needs, let Conv be the output function of the Conv layer. This output function is defined w.r.t. the graph labeling functions:
	
	\begin{subequations}
		\begin{equation}
		B(W',X) = [\mu_I, \zeta_I] \quad\text{where X is a training example}
		%
		%
		\end{equation}
		\begin{equation}
		\label{gcnn:eq:outp}
		\text{Conv}(W,[\mu_I, \zeta_I]) = [\mu, \zeta]
		\end{equation}
		\begin{equation}
		A([\mu, \zeta]) = \hat{Y} \quad\text{s.t. } \hat{Y} \in \mathbb{R}
		\end{equation}
		\begin{equation}
		\text{with } [\mu_I, \zeta_I] = [\mu_I(i)\quad \forall i \in V_I \quad\cdots\quad \zeta_I(ij)\quad \forall ij \in E_I]
		\end{equation}
		\begin{equation}
		\text{and } [\mu, \zeta] = [\mu(i)\quad \forall i \in V_I \quad\cdots\quad \zeta(ij)\quad \forall ij \in E_I]
		\end{equation}
	\end{subequations}
	%
	%
	
	The error gradient for $W$ is calculated using chain derivative:
	
	%
	%
	\begin{subequations}
		\begin{align}
		\frac{\partial\mathcal{J}}{\partial W} &= \frac{\partial A \circ \text{Conv} \circ B}{\partial W}\\
		&= \frac{\partial A}{\partial \text{Conv}} \times \frac{\partial \text{Conv}}{\partial W} &{\scriptstyle \left( \frac{\partial\mathcal{J}}{\partial W} \text{ doesn't depend on }\frac{\partial B}{\partial W} \right)}\\
		&= \frac{\partial \hat{Y}}{\partial [\mu, \zeta]} \times \frac{\partial [\mu, \zeta]}{\partial W}
		\end{align}
	\end{subequations}
	
	Let's assume $\frac{\partial \hat{Y}}{\partial [\mu, \zeta]}$ exists and is known. Therefore, only $\frac{\partial [\mu, \zeta]}{\partial W}$ is to be calculated.
	
	\begin{subequations}
		%
		%
		First of all, we need to expand $[\mu, \zeta]$. Let's expand $\mu$ first for a given vertex $i$ $\in$ $G_I$:
		\begin{align}
		\mu(i) &= \text{GMS}(g_I^i, G_F^W)\\
		&= \left(\underset{y} {\mathrm{max}} \quad s(g_I^i, G_F^W, y)\right)\\
		&= s(g_I^i, G_F^W, \hat{y})\quad\text{with }\hat{y} = \underset{y} {\mathrm{argmax}}\quad s(g_I^i,G_F^W,y)\label{eq:rew}\\
		&= \sum_{k \in V_I^i} s_V(k, \pi(g_I^i, G_F^W,k))  + \sum_{kl \in E_I^i} s_E(kl, \pi(G_I^i, G_F^W, kl))\\
		\mu(i)&=\sum_{k \in V_I^i} \mu_I(k).W_{\pi(g_I^i, G_F, k)}  + \sum_{kl \in E_I^i} \zeta_I(kl).W_{\pi(g_I^i, G_F, kl)}
		\end{align}
		
		Then let's expand $\zeta$:
		\begin{align}
		\zeta(ij) &= \text{score}(ij, G_I, G_F^W)\\
		\zeta(ij) &= \theta \left( \{s_E(ij, \pi(g_I,G_F^W,ij)) \quad \forall g_I \in \Omega_{ij}\} \right)
		\end{align}
		
		If $\theta$ is $\max$, the same rewriting as in Equation~\ref{eq:rew} applies:
		\begin{align}
		\zeta(ij) &= s_E(ij, \pi(g_I^*,G_F^W, ij))\\
		&\text{with } g_I^* = \underset{g_I \in \Omega_{ij}} {\mathrm{argmax}}\quad s_E(ij, \pi(g_I, G_F^W, ij))\\
		\zeta(ij) &= \zeta_I(ij). W_{\pi(g_I^*,G_F^W, ij)}
		\end{align}
		If $\theta$ is avg:
		\begin{align}
		\zeta(ij) &= \frac{1}{|\Omega_{ij}|} \left(\sum_{g_I \in \Omega_{ij}}{\zeta_I(ij). W_{\pi(g_I,G_F^W, ij)}}\right)
		\end{align} 
		
	\end{subequations}
	
	Let's differentiate $\mu$ with respect to $W$. $\forall i \in V_I$:
	\begin{subequations}
		\begin{align} 
		\frac{\partial \mu(i)}{\partial W_a} &= \mu_I(\pi(G_F^W, g_I^i, a)) \quad & \forall a \in V_F\\ 
		\frac{\partial \mu(i)}{\partial W_{ab}} &= \zeta_I(\pi(G_F^W, g_I^i, ab)) &\quad \forall ab \in E_F
		\end{align}
	\end{subequations}
	
	Now let's differentiate $\zeta$. $\forall ij \in E_I$ and $\forall ab \in E_F$:
	
	If $\theta$ is $\max$:
	\begin{subequations}
		\begin{equation}
		\frac{\partial \zeta(ij)}{\partial W_{ab}} = \zeta_I(\pi(g_I^*,G_F^W, ab))
		\end{equation}
	\end{subequations}
	
	If $\theta$ is avg:
	\begin{subequations}
		\begin{align}
		\frac{\partial \zeta(ij)}{\partial W_{ab}} &= \frac{1}{|\Omega_{ij}|} \sum_{g_I \in \Omega_{ij}}{\zeta_I(\pi(g_I,G_F^W, ab))}
		\end{align}
	\end{subequations}
	
	In any case:
	\begin{subequations}
		\begin{equation}
		\label{eq:null}
		\frac{\partial\zeta(ij)}{\partial W_a} = 0 \quad \forall a \in V_I
		\end{equation}
	\end{subequations}
	
	Now, $\frac{\partial\mathcal{J}}{\partial W}$ can be calculated:
	\begin{subequations}
		\begin{align}
		\frac{\partial\mathcal{J}}{\partial W_a} &= \frac{\partial \hat{Y}}{\partial [\mu, \zeta]} \times \frac{\partial [\mu, \zeta]}{\partial W_a} & \forall a \in V_F\\
		&= \sum_{i \in V_I} \frac{\partial \hat{Y}}{\partial \mu(i)}\frac{\partial\mu(i)}{\partial W_a} + \sum_{ij \in E_I} \frac{\partial \hat{Y}}{\partial \zeta(ij)} \frac{\partial\zeta(ij)}{\partial W_a}\\
		&= \sum_{i \in V_I} \frac{\partial \hat{Y}}{\partial \mu(i)}\frac{\partial\mu(i)}{\partial W_a} & \text{w.r.t. Eq~\ref{eq:null}}
		\end{align}
	\end{subequations}
	\begin{subequations}
		\begin{align}
		\frac{\partial\mathcal{J}}{\partial W_{ab}} &= \frac{\partial \hat{Y}}{\partial [\mu, \zeta]} \times \frac{\partial [\mu, \zeta]}{\partial W_{ab}} & \forall ab \in E_F\\
		&= \sum_{i \in V_I} \frac{\partial\mathcal{J}}{\partial \mu(i)}\frac{\partial\zeta(i)}{\partial W_{ab}} + \sum_{ij \in E_I} \frac{\partial\mathcal{J}}{\partial \zeta(ij)} \frac{\partial\zeta(ij)}{\partial W_{ab}}
		\end{align}
	\end{subequations}
	
	%
	%
	%
	Finally, let's suppose $B$ is parameterized with vector $W'$. In this case, $\frac{\partial\mathcal{J}}{\partial W'}$ is to be calculated:
	\begin{subequations}
		\begin{align}
		\frac{\partial\mathcal{J}}{\partial W'} &= \frac{\partial B \circ \text{Conv} \circ A}{\partial W'}\\
		&= \frac{\partial A}{\partial \text{Conv}} \times \frac{\partial \text{Conv}}{\partial B} \times \frac{\partial B}{\partial W'}\\
		&= \frac{\partial \hat{Y}}{\partial [\mu, \zeta]} \times \frac{\partial [\mu, \zeta]}{\partial [\mu_I, \zeta_I]} \times \frac{\partial [\mu_I, \zeta_I]}{\partial W'}
		\end{align}
	\end{subequations}
	
	Let's assume $\frac{\partial [\mu_I, \zeta_I]}{\partial W'}$ exists and is known. $\frac{\partial \hat{Y}}{\partial [\mu, \zeta]}$ has already been evaluated. Therefore, only $\frac{\partial [\mu, \zeta]}{\partial [\mu_I, \zeta_I]}$ is to be calculated. $\forall i \in V_I$:
	\begin{subequations}
		\begin{align}
		\frac{\partial \mu(i)}{\partial \mu_I(k)}
		&=  \begin{cases} 
		W_{\pi(g_I^i, G_F^W, k)} & \text{if } k \in V_I^i \\
		0 & \text{else}
		\end{cases}\quad& \forall k \in V_I\\
		\frac{\partial \mu(i)}{\partial \zeta_I(kl)}
		&=  \begin{cases} 
		W_{\pi(g_I^i, G_F^W, kl)} & \text{if } k \in V_I^i \\
		0 & \text{else}
		\end{cases}\quad& \forall kl \in E_I
		\end{align}
	\end{subequations}
	
	If $\theta$ is $\max$, $\forall kl \in E_I, \forall ij \in E_I$:
	\begin{subequations}
		\begin{equation}
		\frac{\partial \zeta(ij)}{\partial \zeta_I(kl)} = \delta_k^i \delta_j^l W_{\pi(g_I^*,G_F^W, kl)}
		\end{equation}
	\end{subequations}
	
	If $\theta$ is avg, $\forall kl \in E_I, \forall ij \in E_I$:
	\begin{subequations}
		\begin{align}
		\frac{\partial \zeta(ij)}{\partial \zeta_I(kl)} &=
		\delta_k^i \delta_j^l \frac{1}{|\Omega_{ij}|} \sum_{g_I \in \Omega_{ij}}{W_{\pi(g_I,G_F^W, ij)}}
		\end{align}
	\end{subequations}
	
	In any cases:
	\begin{equation}
	\frac{\partial \zeta(ij)}{\partial \mu_I(k)} = 0 \quad\forall k \in V_I
	\end{equation}
	
	%
	
\end{definition}

\subsection{About graph matching differentiation}
In Definition~\ref{def:gcnn:differentiation}, a differentiation of the convolution operator is proposed. This differentiation does not take into account the dependencies between the optimal graph matching $\hat{y}$ and the variables $\{ \mu_I(k) \}_{k \in E_I}$ and $\{ \zeta_I(kl)\}_{kl \in V_I}$. As these variables are used to calculate the possible matchings, it is trivial to conclude such dependencies exist. Nevertheless, the matching solver in use (see Subsection~\ref{subsec:gmcomplexity}) is not differentiable, at least a priori. We therefore assumed $\hat{y}$ as a constant in the gradient calculus with respect to these variables by means of change of variable in Eq.~\ref{eq:rew}.

\subsection{A "no edge matching" version of the graph convolution layer}
This section presents a degraded model. It ignores topology at a local level by not matching edges. It therefore reduces the graph matching problem to a node assignment problem inside a given neighborhood. One concern on this simplification could be that we do not take advantage of the graphs topology. However, topology information is used when computing vertices neighbourhoods. Additionally, this model has lower time complexity as edge information is not taken into account (see details in Subsection~\ref{subsec:gmcomplexity}.)

Used graphs are 3-uplets $(V, E, \mu)$ and the similarity function is simplified as follows:

\begin{subequations}
	\begin{equation}
	s(G_1,G_2,y) =	\sum_{y_{ia}=1} s_V(i, a)
	\end{equation}
	\begin{equation}
	s_V(i, a) = \mu_1(i) . \mu_2(a)
	\end{equation}
\end{subequations}

As a consequence of the edge attributes deletion in the filter graph, its parameter becomes vector $W \in \mathbb{R}^{|V|}$ (as many parameters as vertex). The filter is defined as follows:

\begin{subequations}
	\begin{equation}
	G_F^W = (V_F, E_F, \mu_F^W)
	\end{equation}	
	\begin{align}
	\text{with}\quad
	& \mu_F^W(a) = W_a
	\end{align}
\end{subequations}

The output function of the filter $u : \mathbb{G} \to \mathbb{G}$ is defined as follows:
\begin{subequations}
	\begin{align}
	u(G_I) &= G_I \odot G_F^W\\
	&= (V_I, E_I, \mu)
	\end{align}
\end{subequations}

\subsection{Graph pooling}
As in euclidean convolutional neural nets, we want to implement not only convolutional layers but also pooling/downsampling layers. In the existing literature, downsampling is view as graph coarsening \cite{DBLP:journals/corr/BronsteinBLSV16}. A recurrent graph coarsening algorithm choice seems to be Graclus \cite{dhillon2007weighted} (used in \cite{DBLP:journals/corr/MontiBMRSB16,DBLP:journals/corr/DefferrardBV16}).

We propose to use a community detection algorithm (Louvain method \cite{2008JSMTE..10..008B}) as the base of our graph pooling layer. Louvain method deals with weighted graphs. In our case, edge weights are computed by scalar products of involved vertices. This choice is brought by the following intuition: the higher nodes attributes scalar product get, the more these vertices probabilities to fall in the same cluster increases (because a higher scalar product implies vector similarity).

\subsection{Hyperparameterization}
\label{gcnn:sec:hyperparam}

As in any neural network, graph neural networks have parameters that won't be optimized from gradient descent.

The first one is the graph filter (its number of nodes and adjacency matrix). The number of nodes in the graph filter is analogous to the size of a classic convolution kernel. A $3 \times 3$ kernel filter is equivalent to a 9 nodes filter graph with grid-like adjacency. The second hyperparameter is the size of extracted neighbourhoods graphs which is the maximum node distance in a given node neighbourhood. A 2-hop-sized neighbourhoods will include nodes that can be reached from the origin node in two hops or less.

These hyperparameters could be optimized through grid or random search. However, to restrain our study, we will consider the following postulate: a graph filter should be congruent with extracted neighbourhoods. In other words, the two should have equal sizes and identical topologies as much as possible. This postulate comes from classic graph convolution where each kernel coefficient is matched with one and only one image coefficient.

\subsection{Choosing the graph matching solver}
\label{subsec:gmcomplexity}
\newcommand*\mean[1]{\overline{#1}}
The algorithm for solving the graph matching problem is a critical element for the model. The first reason is that it is potentially the highest in complexity since graph matching problems are up to NP-hard. Additionally, graph matching is solved as many times as there are vertices in the input graph (the size of every problem to solve being that of every vertex neighbourhood).

We opted for a bipartite (BP) graph matching algorithm \cite{DBLP:journals/ivc/RiesenB09}. Complexity of such an algorithm is among the lowest (polynomial time) for solving error-tolerant graph matching problems suboptimally. 

Bipartite graph matching algorithm reduces graph matching to vertex matching by embedding an estimation for edge costs in the vertex costs. This edge cost estimation is computed by solving an edge-assignment problem for every node-matching possibility. Therefore, BP has to solve as many matching problems as there are edge-costs.

We used a variant of BP called Square Fast BP \cite{serratosa2015speeding} where the cost matrix for vertex matching is of size $\max(|g_I|, |G_F|) \times \max(|g_I|, |G_F|)$ with $|G_F|$ and $|g_I|$ being number of vertices in filter graph $G_F$ and neighbourhood graph $g_I$. Assuming both neighbourhood and filter graphs are complete, a matching problem complexity is $O(\max(|g_I|, |G_F|)^3)$.

As a consequence, worst case complexity with fast bipartite matching is the following:
$$O(\max(|g_I|, |G_F|)^5)$$

Some preliminary experiments showed impracticable computation time of the full model. 
As a first workaround, the experimental part of this paper will focus on "no edge matching" model. This workaround allowed to keep processing to an acceptable level (that is suitable for small classification experiments). Edge cost estimation by edge matching is no longer required. The simplified model has the following pointwise complexity:
$$O(\max(|g_I|, |G_F|)^3)$$

\section{Experimental work}
\label{gcnn:sec:experiment}
In this section, we test the model according to several parameters. We want to test our model with a simple classification task on MNIST digit images.

\subsection{Baselines}
 Our approach was compared with two other approaches:

\begin{itemize}
	\item Vanilla CNN layer
	\item \cite{DBLP:journals/corr/MontiBMRSB16} mixture model graph CNN.
\end{itemize}

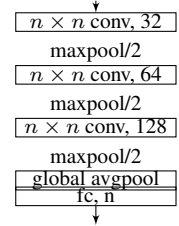
\begin{figure}[h]
	\centering
	\begin{tikzpicture}[auto, >=latex',font=\footnotesize]
	
	\node [pooling, text width=2cm] (input) {};
	\node [block, afterpool, below of=input] (conv1) {$n\times n$ conv, 32};
	\node [pooling, below of=conv1] (pool1) {maxpool/2};
	\node [block, afterpool, below of=pool1] (conv2) {$n\times n$ conv, 64};
	\node [pooling, below of=conv2] (pool2) {maxpool/2};
	\node [block, afterpool, below of=pool2] (conv3) {$n\times n$ conv, 128};
	\node [pooling, below of=conv3] (pool3) {maxpool/2};
	\node [block, afterpool, below of=pool3] (flatten) {global avgpool};
	
	\node [block, below of=flatten] (fc1) {fc, n};
	
	\node [below of=fc1,pooling] (output) {};
	
	\draw [->] (input) -- (conv1);
	\draw [->] (fc1) -- (output);
	
	\end{tikzpicture}
	
	\caption{Network structure used for graph convolution experiments}
	\label{gcnn:fig:network}
\end{figure}

Same network topology was used for all approaches. It consists of classical ConvPool blocks linearly connected. Figure~\ref{gcnn:fig:network} shows the exact network structure in use. In case of graph convolution, $n\times n$ convolution filters equivalents are $n^2$ nodes filters and $2 \times 2$ pooling becomes 4 nodes pooling. $n$ is set depending on average graph connectivity in a given dataset: if the average number of neighbours in a given dataset is 9, $n=9$.

The last layer is a global pooling one. As in the euclidean case, it consists in aggregating each filter feature map in one scalar value. In our case, feature maps are aggregated by taking its average value.

\subsection{Data}
Quantitative experiments in this section are operated on digit images of MNIST dataset \cite{lecun1998gradient}. We chose this dataset as this was in use in the graph convolution literature.  MNIST is a good "hello world" machine learning (ML) dataset. MNIST helps at quickly iterating on the learning model. Performance information gathered from experiments on MNIST can be great for judging how the model might perform on much harder and larger datasets like ImageNet.

In addition to the original MNIST dataset, a rotated version was used \cite{larochelle2007empirical}. To compare results with MNIST-rotated, MNIST-original has to be modified as follows. MNIST-reduced proportions are unusual: 10000, 2000 and 50000 images respectively for train, validation and test whereas MNIST-original has 60000 and 10000 images respectively for train/validation and test. We used MNIST-reduced, a resampled version of MNIST-original to fit MNIST-rotated ratio between subsets cardinalities: MNIST-reduced and MNIST-rotated have both 10000, 2000 and 50000 images respectively for train, validation and test. All the set cardinalities are summed up in Table~\ref{gcnn:tab:sets}. Note that the test set of MNIST-reduced is larger than the training set by a factor 5 consequently, the generalization ability is better assessed.

\begin{table}[ht!]
	\small
	\centering
	\caption{Different MNIST-based graph datasets}
	\label{gcnn:tab:sets}
	\begin{tabular}{| l || c | c | c |}
		\hline
		\textbf{Dataset} & \textbf{Training set} & \textbf{Validation set} & \textbf{Testing set} \\
		\hline
		MNIST-original & 48~000 & 12~000 & 10~000 \\
		\hline
		MNIST-rotated & \multirow{3}{*}{10~000} & \multirow{3}{*}{2~000} & \multirow{3}{*}{50~000} \\
		\cline{1-1}
		MNIST-reduced &  &  &  \\
		\cline{1-1}
		MNIST-mixed &  &  &  \\
		\hline
	\end{tabular}
\end{table}

Lastly, to test rotation invariance, a third MNIST-based dataset was added: MNIST-mixed. It was generated by combining MNIST-reduced train and validation sets and MNIST-rotated test set. It is design so that the models are trained on rotation-free images but tested on rotated images.

\begin{figure}[ht!]
	\centering
    \begin{tabular}{cccc}
         \includegraphics[width=0.1\textwidth]{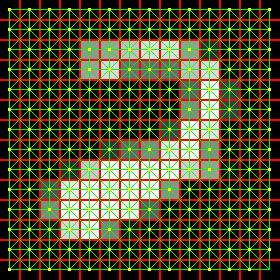} &
         \includegraphics[width=0.1\textwidth]{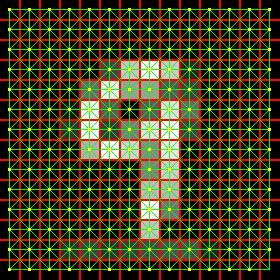} &
         \includegraphics[width=0.1\textwidth]{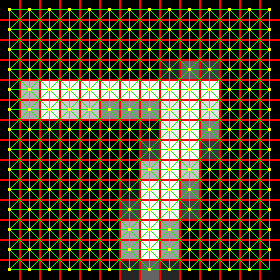} &
         \includegraphics[width=0.1\textwidth]{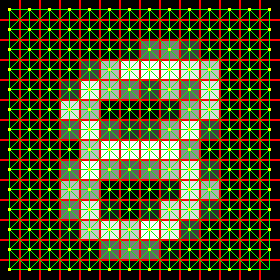} \\
         \includegraphics[width=0.1\textwidth]{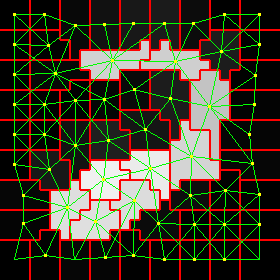} &
         \includegraphics[width=0.1\textwidth]{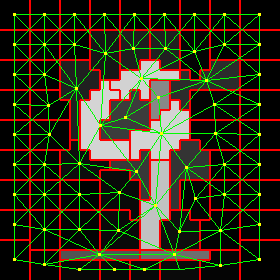} &
         \includegraphics[width=0.1\textwidth]{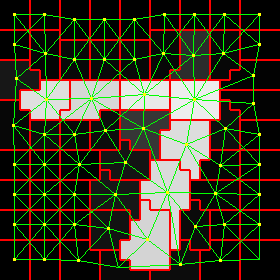} &
         \includegraphics[width=0.1\textwidth]{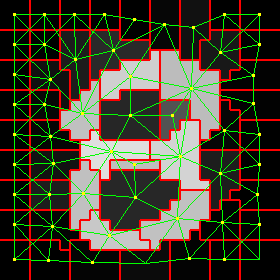}
    \end{tabular}
    \caption{MNIST graphs. Top is $\frac{1}{4}$ grid, bottom is 75 superpixels RAG. Red symbolizes vertex frontiers and green shows edges.}
	\label{gcnn:fig:graph_samples}
\end{figure}

As MNIST is an image dataset, a graph-based representation of images has to be chosen. Representations used in \cite{DBLP:journals/corr/MontiBMRSB16} are superpixels graphs and grid graphs. We used $\frac{1}{4}$ grids ($28\times28$ images resized to $14\times14$) and generated 75 superpixels Region Adjacency Graphs (RAG) using SLIC algorithm \cite{achanta2012slic} with superpixel adjacency as edges (see Table~\ref{gcnn:tab:representations}). Sample graphs are depicted in Figure~\ref{gcnn:fig:graph_samples}.

\begin{table*}
	\centering
	\caption{MNIST representations}
	\label{gcnn:tab:representations}
	\begin{tabular}{|p{2.2cm} || c | c | c | }
		\hline
		\textbf{Representation} & \textbf{Nb nodes}  & \textbf{Vertex attributes} & \textbf{Edge attributes}\\
		\hline
		$\frac{1}{4}$ grid & $14^2$ &  Pixel intensities & \multirow{2}{*}{Relative polar coordinates}\\
		\cline{1-3}
		75 superpixels & 75 (average) & Average superpixel intensities & \\
		\hline
	\end{tabular}
\end{table*}

\subsection{Parameterization}
Following hyperparameters were set after preliminary tests were conducted: Models are trained during 50 epochs using Adaptive Moment (Adam) gradient descent (learning rate $10^{-3}$). Neighbourhood reach in use is 1-hop and filter size was set in accordance with average neighbourhood size (9 nodes).

\subsection{Protocol}
Following experiments were conducted:
\begin{description}
	\item[Experiment 1] Models are tested on MNIST digit images classification task
	\item[Experiment 2] Several neighbourhood connectivities are tested on our model (1 and 2 hops)
	\item[Experiment 3] Rotation invariance is investigated. Spatial information for our datasets is conveyed by edge attributes. In such a frame, as our "no edges" model ignores edge attributes, it is theoretically rotation-invariant. Experiment~3 aims at experimentally validating this claim. This is done by training models on unrotated images and testing on rotated ones. MNIST-mixed set is used to this end. 
	\item[Experiment 4] A sample filter is visualized on some MNIST example images
	\item[Experiment 5] Graph based methods are tested on regular grids and on irregular graphs (75 superpixels RAG) for testing sensitivity to domain changes
\end{description}

As stated before and because of technical limitations, experiments involving MNIST datasets will focus on the two first MNIST classes (referred to as MNIST-2class)

\subsection{Results}
Results on MNIST-2class are listed on Table~\ref{gcnn:tab:results}. Results include classification from both $\frac{1}{4}$ grid graphs and SLIC 75-superpixels graphs. 
This table shows results for each dataset using classic CNN, MoNet \cite{DBLP:journals/corr/MontiBMRSB16} and our method.

\subsubsection{Experiment 1: MNIST}

On MNIST-2class, our model competes in a 3\% margin with used baselines.

\subsubsection{Experiment 2: Neighbourhood size}
Extending the neighbourhood size did not have any significant effect on performance (see Table~\ref{gcnn:tab:neighbourhood})

\subsubsection{Experiment 3: Rotation invariance}
\label{subsec:exp3}

On MNIST-mixed, no performance loss was observed on testing for our method. This is especially visible on grid graphs results where only classic CNN and MoNet show a 10~percent loss. A trivial explanation of how is this invariance obtained is that our graph convolution filters are non-oriented because edge attributes are ignored.

\begin{table*}
	\centering
	\caption{Recognition rates on MNIST 2class}
	\label{gcnn:tab:results}
	\begin{tabular}{| p{3cm} l | c c | c c | c c |}
		\hline
		\textbf{Representation} & \textbf{Dataset} & \multicolumn{2}{c|}{\textbf{CNN}} & \multicolumn{2}{c|}{\textbf{MoNet}} & \multicolumn{2}{c|}{\textbf{Ours}}\\
		& & Valid & Test & Valid & Test & Valid & Test \\
		\hline
		\multirow{2}{*}{$\frac{1}{4}$ grid} & MNIST reduced & 100 \% & 99.88 \% & 97.56 \% & 99.40 \% & 99.51 \% & 97.76 \% \\
		& MNIST mixed &  100 \% & 89.87 \% & 97.76 \%  & 88.90 \% & 99.27 \% & 95.63 \% \\
		\hline
		\multirow{2}{*}{75 superpixels} & MNIST reduced & & & 94.13 \% & 92.70 \% & 94.13 \% & 89.53 \% \\
		& MNIST mixed & & & 94.13 \% & 92.90 \% & 94.62 \% & 94.17 \% \\
		\hline
	\end{tabular}
\end{table*}

\begin{table}
	\small
	\centering
	\caption{Recognition rates for different neighbourhood sizes on MNIST reduced 2 class}
	\label{gcnn:tab:neighbourhood}
	\begin{tabular}{|c | c c | c c |}
		\hline
		\textbf{Representation} & \multicolumn{2}{c|}{1 hop} & \multicolumn{2}{c|}{2 hops}\\
		& Valid & Test & Valid & Test\\
		\hline
		$\frac{1}{4}$ grid & 99.02\% & 97.55\% & 98.04\% & 96.47\% \\
		\hline
		75 superpixels & 97.55\% & 93.74\% & 96.82\% & 93.62\% \\
		\hline
	\end{tabular}
\end{table}

\subsubsection{Experiment 4: Visualizing graph convolution on images}
\label{subsec:exp4}

As an additional experimental material, we tried to visualize the result of a handcrafted filter on images. As for euclidean convolution, the most straightforward filter operation is edge detection. This is usually done by using Sobel operator that calculates intensity gradient at each spatial point of the image. 

A potential equivalent graph convolution filter is $\left(-1 \hspace{0.2cm} 1\right)$ (the filter is a 2-nodes graph with respective attributes $-1$ and $1$.) The intuition behind this filter is that the nodes will be matched respectively to the lowest (for the attributed $-1$ node) and highest (for the attributed $1$ node) intensities. As a consequence, this filter will find the highest node attribute difference in every node neighbourhood, making it a sort of eager edge detection filter.

\begin{figure}[ht!]
	\centering
	\begin{tabular}{ccc}
	     \includegraphics[width=0.1\textwidth]{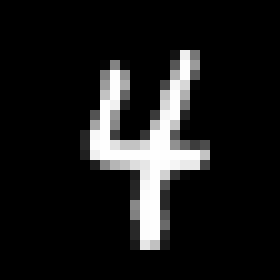} &
	     \includegraphics[width=0.1\textwidth]{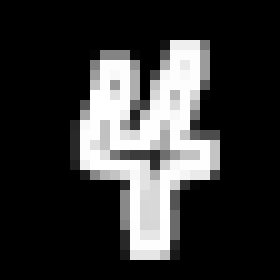} &
	     \includegraphics[width=0.1\textwidth]{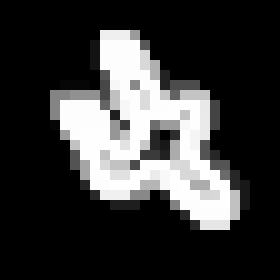}\\
	     \includegraphics[width=0.1\textwidth]{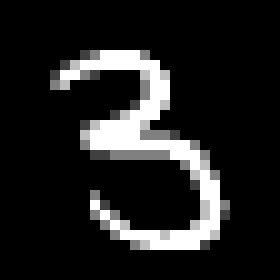} &
	     \includegraphics[width=0.1\textwidth]{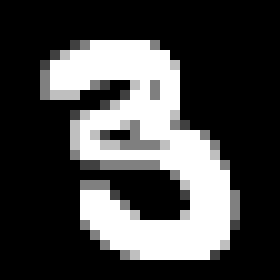} &
	     \includegraphics[width=0.1\textwidth]{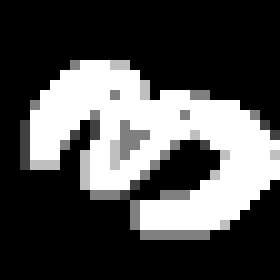} \\
	     \includegraphics[width=0.1\textwidth]{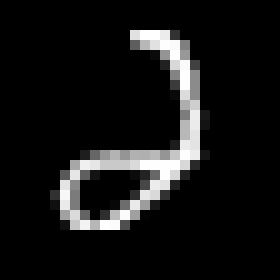} &
	     \includegraphics[width=0.1\textwidth]{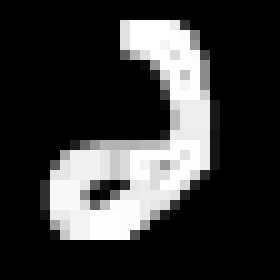} &
	     \includegraphics[width=0.1\textwidth]{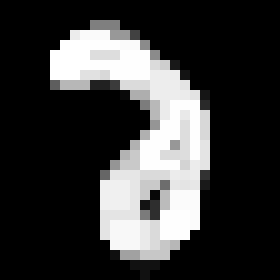} \\
	     \includegraphics[width=0.1\textwidth]{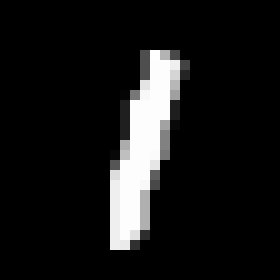} &
	     \includegraphics[width=0.1\textwidth]{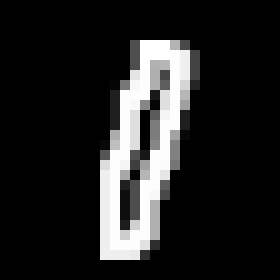} &
	     \includegraphics[width=0.1\textwidth]{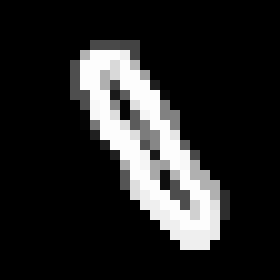}
	\end{tabular}
	\caption{MNIST graph convolution examples (respectively original, convoluted and rotated convoluted versions)}
	\label{gcnn:fig:image_samples}
\end{figure}

We applied this filter on grid graphs to visualize the output graph as an image (as the graph-to-image transformation is trivial). Figure~\ref{gcnn:fig:image_samples} shows example applications of this filter on both original and rotated examples. This last figure suggests rotation invariance.

\subsubsection{Experiment 5: Testing graph convolution across domain}
\label{subsec:exp5}

A particular concern on graph convolution operators is sensitivity to domain changes, i.e. capacity to identify similarities on irregular graphs. Both graph convolution tested show little performance loss between regular (grids) and irregular (75 superpixels RAG) results.

\subsubsection{Training duration}
\label{subsec:duration}
As mentioned in Subsection~\ref{subsec:gmcomplexity}, complexity of the model makes experiment tedious to lead. Epoch durations are given in Table~\ref{gcnn:tab:epochdur}.

\begin{table}
	\centering
	\caption{Epoch durations on MNIST 2class (Models use different implementations/hardware: CNN is Keras on GPU, MoNet is Theano on GPU and Ours is Keras on CPU)}
	\label{gcnn:tab:epochdur}
	\begin{tabular}{| l | l | l | l |}
		\hline
		\textbf{Representation} & \textbf{CNN} & \textbf{MoNet} & \textbf{Ours}\\
		\hline
		$\frac{1}{4}$ grid & 1s & 1s & 17min 29s \\
		\hline
		75 superpixels & NA & 1s & 2min 42s \\
		\hline
	\end{tabular}
\end{table}

\section{Conclusion and perspectives}
\label{sec:conclusion}
In this paper, a graph convolutional neural network layer is proposed and tested in a simplified form.

Our model performance is at state of the art level on simple tasks. It shows robustness with respect to graph domain changes.

Following improvements could highly benefit to performances and computational costs. The bipartite solver is not the most suitable choice for our use. Complexity seems to be too high for an efficient application. Using a less complex solver would allow the full model to be used in practice and applied to larger graphs. Using the edge information would probably enhance performances significantly. Moreover, it will probably help with solving more complex problems. 

Another point of improvement is regarding differentiation: the solver operator is not differentiable. The gradient must then be approximated by neglecting contribution of the solver intermediary states. Finding a differentiable solver would enhance trainability of the model.

Addressing these issues will not only enhance the current degraded version of the model but also allow to implement the full model in a usable form. This model has the peculiarity to learn edge attributes as well as vertex attributes. It is to our knowledge the only graph convolution formulation that suggests to modify the spatiality of edge attibutes.

Finally, investigating our downsampling layer would justify a whole study for itself. It would be interesting to study the quality of the downsampled graphs but also to study the effect of weighting edges regarding vertex similarity.

\section{Code}
Code for running the model can be found at \url{https://github.com/prafiny/graphconv}

\bibliographystyle{plain}
\bibliography{ms}

\begin{thebibliography}{10}

\bibitem{achanta2012slic}
Radhakrishna Achanta, Appu Shaji, Kevin Smith, Aurelien Lucchi, Pascal Fua, and
  Sabine S{\"u}sstrunk.
\newblock Slic superpixels compared to state-of-the-art superpixel methods.
\newblock {\em IEEE transactions on pattern analysis and machine intelligence},
  34(11):2274--2282, 2012.

\bibitem{2008JSMTE..10..008B}
Vincent~D. {Blondel}, Jean-Loup {Guillaume}, Renaud {Lambiotte}, and Etienne
  {Lefebvre}.
\newblock {Fast unfolding of communities in large networks}.
\newblock {\em Journal of Statistical Mechanics: Theory and Experiment},
  2008(10):10008, Oct 2008.

\bibitem{DBLP:journals/prl/BougleuxBCFGV17}
S{\'{e}}bastien Bougleux, Luc Brun, Vincenzo Carletti, Pasquale Foggia, Benoit
  Ga{\"{u}}z{\`{e}}re, and Mario Vento.
\newblock Graph edit distance as a quadratic assignment problem.
\newblock {\em Pattern Recognition Letters}, 87:38--46, 2017.

\bibitem{DBLP:journals/corr/BronsteinBLSV16}
Michael~M. Bronstein, Joan Bruna, Yann LeCun, Arthur Szlam, and Pierre
  Vandergheynst.
\newblock Geometric deep learning: going beyond euclidean data.
\newblock {\em CoRR}, abs/1611.08097, 2016.

\bibitem{DBLP:conf/sspr/BunkeR08}
Horst Bunke and Kaspar Riesen.
\newblock Graph classification on dissimilarity space embedding.
\newblock In Niels da~Vitoria~Lobo, Takis Kasparis, Fabio Roli, James~Tin{-}Yau
  Kwok, Michael Georgiopoulos, Georgios~C. Anagnostopoulos, and Marco Loog,
  editors, {\em Structural, Syntactic, and Statistical Pattern Recognition,
  Joint {IAPR} International Workshop, {SSPR} {\&} {SPR} 2008, Orlando, USA,
  December 4-6, 2008. Proceedings}, volume 5342 of {\em Lecture Notes in
  Computer Science}, page~2. Springer, 2008.

\bibitem{DBLP:journals/corr/DefferrardBV16}
Micha{\"{e}}l Defferrard, Xavier Bresson, and Pierre Vandergheynst.
\newblock Convolutional neural networks on graphs with fast localized spectral
  filtering.
\newblock {\em CoRR}, abs/1606.09375, 2016.

\bibitem{dhillon2007weighted}
Inderjit~S Dhillon, Yuqiang Guan, and Brian Kulis.
\newblock Weighted graph cuts without eigenvectors a multilevel approach.
\newblock {\em IEEE transactions on pattern analysis and machine intelligence},
  29(11):1944--1957, 2007.

\bibitem{gauzere}
Benoit Ga\"{u}zere, Luc Brun, and Didier Villemin.
\newblock Two new graphs kernels in chemoinformatics.
\newblock {\em Pattern Recogn. Lett.}, 33(15):2038 -- 2047, 2012.

\bibitem{DBLP:journals/corr/HamiltonYL17}
William~L. Hamilton, Rex Ying, and Jure Leskovec.
\newblock Inductive representation learning on large graphs.
\newblock {\em CoRR}, abs/1706.02216, 2017.

\bibitem{DBLP:journals/corr/KipfW16}
Thomas~N. Kipf and Max Welling.
\newblock Semi-supervised classification with graph convolutional networks.
\newblock {\em CoRR}, abs/1609.02907, 2016.

\bibitem{larochelle2007empirical}
Hugo Larochelle, Dumitru Erhan, Aaron Courville, James Bergstra, and Yoshua
  Bengio.
\newblock An empirical evaluation of deep architectures on problems with many
  factors of variation.
\newblock In {\em Proceedings of the 24th international conference on Machine
  learning}, pages 473--480. ACM, 2007.

\bibitem{lecun1998gradient}
Yann LeCun, L{\'e}on Bottou, Yoshua Bengio, Patrick Haffner, et~al.
\newblock Gradient-based learning applied to document recognition.
\newblock {\em Proceedings of the IEEE}, 86(11):2278--2324, 1998.

\bibitem{Leordeanu2009}
Marius Leordeanu, Martial {Hebert }, and Rahul Sukthankar.
\newblock An integer projected fixed point method for graph matching and map
  inference.
\newblock In {\em Proceedings Neural Information Processing Systems}, pages
  1114--1122, 2009.

\bibitem{LiuQ14}
Zhiyong Liu and Hong Qiao.
\newblock {GNCCP} - graduated nonconvexityand concavity procedure.
\newblock {\em {IEEE} Trans. Pattern Anal. Mach. Intell.}, 36:1258--1267, 2014.

\bibitem{DBLP:journals/pr/LuqmanRLB13}
Muhammad~Muzzamil Luqman, Jean{-}Yves Ramel, Josep Llad{\'{o}}s, and Thierry
  Brouard.
\newblock Fuzzy multilevel graph embedding.
\newblock {\em Pattern Recognition}, 46(2):551--565, 2013.

\bibitem{DBLP:journals/corr/MontiBMRSB16}
Federico Monti, Davide Boscaini, Jonathan Masci, Emanuele Rodol{\`{a}}, Jan
  Svoboda, and Michael~M. Bronstein.
\newblock Geometric deep learning on graphs and manifolds using mixture model
  cnns.
\newblock {\em CoRR}, abs/1611.08402, 2016.

\bibitem{DBLP:series/smpai/NeuhausB07}
Michel Neuhaus and Horst Bunke.
\newblock {\em Bridging the Gap between Graph Edit Distance and Kernel
  Machines}, volume~68 of {\em Series in Machine Perception and Artificial
  Intelligence}.
\newblock WorldScientific, 2007.

\bibitem{DBLP:journals/corr/NowakVBB17}
Alex Nowak, Soledad Villar, Afonso~S. Bandeira, and Joan Bruna.
\newblock A note on learning algorithms for quadratic assignment with graph
  neural networks.
\newblock {\em CoRR}, abs/1706.07450, 2017.

\bibitem{DBLP:journals/jvcir/RaveauxBO13}
Romain Raveaux, Jean-Christophe Burie, and Jean-Marc Ogier.
\newblock Structured representations in a content based image retrieval
  context.
\newblock {\em J. Visual Communication and Image Representation},
  24(8):1252--1268, 2013.

\bibitem{livreRiesen15}
Kaspar Riesen.
\newblock {\em Structural Pattern Recognition with Graph Edit Distance -
  Approximation Algorithms and Applications}.
\newblock Advances in Computer Vision and Pattern Recognition. Springer, 2015.

\bibitem{DBLP:journals/ivc/RiesenB09}
Kaspar Riesen and Horst Bunke.
\newblock Approximate graph edit distance computation by means of bipartite
  graph matching.
\newblock {\em Image Vision Comput.}, 27(7):950--959, 2009.

\bibitem{DBLP:journals/pami/RothLKB03}
Volker Roth, Julian Laub, Motoaki Kawanabe, and Joachim~M. Buhmann.
\newblock Optimal cluster preserving embedding of nonmetric proximity data.
\newblock {\em {IEEE} Trans. Pattern Anal. Mach. Intell.}, 25(12):1540--1551,
  2003.

\bibitem{serratosa2015speeding}
Francesc Serratosa.
\newblock Speeding up fast bipartite graph matching through a new cost matrix.
\newblock {\em International Journal of Pattern Recognition and Artificial
  Intelligence}, 29(02):1550010, 2015.

\bibitem{2017arXiv171010903V}
Petar {Veli{\v{c}}kovi{\'c}}, Guillem {Cucurull}, Arantxa {Casanova}, Adriana
  {Romero}, Pietro {Li{\`o}}, and Yoshua {Bengio}.
\newblock {Graph Attention Networks}.
\newblock {\em arXiv e-prints}, page arXiv:1710.10903, Oct 2017.

\bibitem{DBLP:journals/corr/abs-1901-00596}
Zonghan Wu, Shirui Pan, Fengwen Chen, Guodong Long, Chengqi Zhang, and
  Philip~S. Yu.
\newblock A comprehensive survey on graph neural networks.
\newblock {\em CoRR}, abs/1901.00596, 2019.

\bibitem{DBLP:journals/corr/abs-1812-04202}
Ziwei Zhang, Peng Cui, and Wenwu Zhu.
\newblock Deep learning on graphs: {A} survey.
\newblock {\em CoRR}, abs/1812.04202, 2018.

\bibitem{DBLP:journals/corr/abs-1812-08434}
Jie Zhou, Ganqu Cui, Zhengyan Zhang, Cheng Yang, Zhiyuan Liu, and Maosong Sun.
\newblock Graph neural networks: {A} review of methods and applications.
\newblock {\em CoRR}, abs/1812.08434, 2018.

\end{thebibliography}

\end{document}